\documentstyle[latex-acl]{article}
\title{\vspace{-0.5in}\LARGE\bf DUAL-CODING THEORY AND CONNECTIONIST LEXICAL
SELECTION}
\author{Ye-Yi Wang\thanks{This work was partly supported
by ARPA and ATR Interpreting Telephony Research Laboratorie.}\\
Computational Linguistics Program\\
Carnegie Mellon University \\
Pittsburgh, PA 15232 \\
Internet: yyw@cs.cmu.edu\\}

\makeatletter

\def\@citex[#1]#2{\if@filesw\immediate\write\@auxout{\string\citation{#2}}\fi
  \def\@citea{}\@cite{\@for\@citeb:=#2\do
    {\@citea\def\@citea{;\penalty\@m\ }\@ifundefined
       {b@\@citeb}{{\bf ?}\@warning
       {Citation `\@citeb' on page \thepage \space undefined}}%
{\csname b@\@citeb\endcsname}}}{#1}}
\let\@internalcite\cite
\def\cite{\def\citename##1{##1, }\@internalcite}
\def\shortcite{\def\citename##1{}\@internalcite}
\def\newcite{\leavevmode\def\citename##1{{##1} (}\@internalciteb}

\def\@citexb[#1]#2{\if@filesw\immediate\write\@auxout{\string\citation{#2}}\fi
  \def\@citea{}\@newcite{\@for\@citeb:=#2\do
    {\@citea\def\@citea{;\penalty\@m\ }\@ifundefined
       {b@\@citeb}{{\bf ?}\@warning
       {Citation `\@citeb' on page \thepage \space undefined}}%
\hbox{\csname b@\@citeb\endcsname}}}{#1}}
\def\@internalciteb{\@ifnextchar
[{\@tempswatrue\@citexb}{\@tempswafalse\@citexb[]}}

\def\@newcite#1#2{{#1\if@tempswa, #2\fi)}}

\def\@biblabel#1{\def\citename##1{##1}[#1]\hfill}

\def\@cite#1#2{({#1\if@tempswa , #2\fi})}

\def\thebibliography#1{\vskip\parskip%
\vskip\baselineskip%
\def\baselinestretch{1}%
\ifx\@currsize\normalsize\@normalsize\else\@currsize\fi%
\vskip-\parskip%
\vskip-\baselineskip%
\section*{References\@mkboth
 {References}{References}}\list
 {}{\setlength{\labelwidth}{0pt}\setlength{\leftmargin}{\parindent}
 \setlength{\itemindent}{-\parindent}}
 \def\newblock{\hskip .11em plus .33em minus -.07em}
 \sloppy\clubpenalty4000\widowpenalty4000
 \sfcode`\.=1000\relax}

\def\thesourcebibliography#1{\vskip\parskip%
\vskip\baselineskip%
\def\baselinestretch{1}%
\ifx\@currsize\normalsize\@normalsize\else\@currsize\fi%
\vskip-\parskip%
\vskip-\baselineskip%
\section*{Sources of Attested Examples\@mkboth
 {Sources of Attested Examples}{Sources of Attested Examples}}\list
 {}{\setlength{\labelwidth}{0pt}\setlength{\leftmargin}{\parindent}
 \setlength{\itemindent}{-\parindent}}
 \def\newblock{\hskip .11em plus .33em minus -.07em}
 \sloppy\clubpenalty4000\widowpenalty4000
 \sfcode`\.=1000\relax}

\def\@lbibitem[#1]#2{\item[]\if@filesw
      { \def\protect##1{\string ##1\space}\immediate
        \write\@auxout{\string\bibcite{#2}{#1}}\fi\ignorespaces}}

\def\@bibitem#1{\item\if@filesw \immediate\write\@auxout
       {\string\bibcite{#1}{\the\c@enumi}}\fi\ignorespaces}

\makeatother

\begin{document}
\input{psfig}

\maketitle
\bibliographystyle{latex-acl}
\vspace{-0.5in}
\begin{abstract}

We introduce the bilingual dual-coding theory as a model for bilingual
mental representation. Based on this model, lexical selection neural
networks are implemented for a connectionist transfer project in machine
translation.

\end{abstract}

\section{Introduction}

Psycholinguistic knowledge would be greatly helpful, as we believe, in
constructing an artificial language processing system. As for machine
translation, we should take advantage of our understandings of (1) how the
languages are represented in human mind; (2) how the representation is
mapped from one language to another; (3) how the representation and mapping
are acquired by human.

The bilingual dual-coding theory \cite{paivio-86} partially answers the
above questions. It depicts the verbal representations for two different
languages as two separate but connected logogen systems, characterizes the
translation process as the activation along the connections between the
logogen systems, and attributes the acquisition of the representation to
some {\em unspecified} statistical processes.

We have explored an information theoretical neural network \cite{gorin-89}
that can acquire the verbal associations in the dual-coding theory.  It
provides a learnable lexical selection sub-system for a connectionist
transfer project in machine translation.

\section{Dual-Coding Theory}

There is a well-known debate in psycholinguistics concerning the bilingual
mental representation: independence position assumes that bilingual memory
is represented by two functionally independent storage and retrieval
systems, whereas interdependence position hypothesizes that all information
of languages exists in a common memory store. Studies on cross-language
transfer and cross-language priming have provided evidence for both
hypotheses \cite{de-groot-91,lambert-58}.

\begin{figure}
\centerline{\psfig{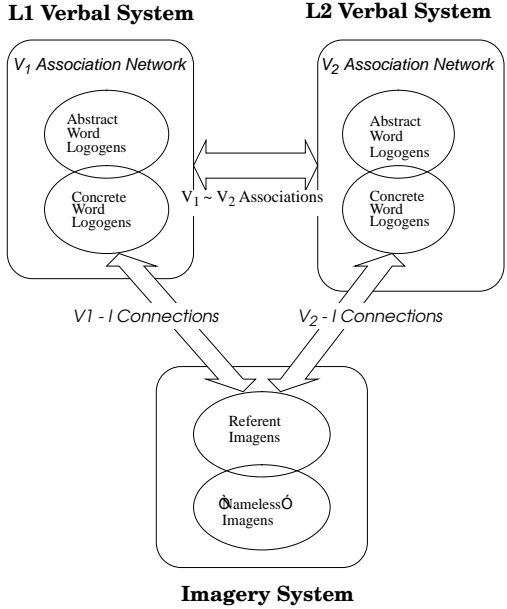}}
\caption{Bilingual Dual-Coding Representation}
\label{dual}
\end{figure}

Dual-coding theory explains the coexistence of independent and
interdependent phenomena with separate but connected structures.  The
general dual-coding theory hypothesizes that human represents language with
dual systems --- the verbal system and the imagery system. The elements of
the verbal system are {\em logogens} for words in a language. The elements
of the imagery system, called ``{\em imagens}'', are connected to the
logogens in the verbal systems via {\it referential connections}. Logogens
in a verbal system are also interconnected with {\it associative
connections}.  The bilingual dual-coding theory proposes an architecture in
which a common imagery system is connected to two verbal systems, and the
two verbal systems are interconnected to each other via associative
connections [Figure \ref{dual}]. Unlike the within-language associations,
which are rich and diverse, these between-language associations involve
primarily translation equivalent terms that are experienced together
frequently.  The interconnections among the three systems explain the
interdependent functional behavior. On the other hand, the different
characteristics of within-language and between-language associations
account for the independent functional behavior.

Based on the above structural assumption, dual-coding theory proposes a
parallel set of processing assumptions. Activation of connections between
referentially related imagens and logogens is called {\it referential
processing}.  Naming objects and imaging to words are prototypical
examples. Activation of associative connections between logogens is called
{\it associative processing}. Lexical translation is an example of
associative processing between two languages.

\section{Connectionist Lexical Selection}

\subsection{Lexical Selection}

Lexical selection is the task of choosing target language words that accurately
reflect the meaning of the corresponding source language words.
It plays an important role in machine translation \cite{pustejovsky-87}.

A common lexical selection practice involves an intermediate representation.
It disambiguates the source language words to entities in the intermediate
representation, then maps from the entities to the target lexical entries. This
intermediate representation may be Lexical Concept Structure \cite{dorr-89}
or interlingua \cite{nirenburg-87}. This engineering approach
requires great effort in designing the representation and the mapping rules.

Currently, there are some efforts in statistical lexical selection. A
target language word {\it W$_{\hbox{t}}$} can be selected with the
posterior probability Pr({\it W$_{\hbox{t}}$}{\tt\char`\|}{\it
W$_{\hbox{s}}$}) given the source language word {\it W$_{\hbox{s}}$}.
Several target language lexical entries may be selected for a single source
language word. Then the correct selections can be identified by the
language model of the target language \cite{brown-90}. This approach is
learnable. However, the accuracy is low. One reason is that it does not use
any structural information of a language.

In next subsections, we propose information-theoretical networks
based on the bilingual dual-coding theory for lexical selection.

\subsection{Information-Theoretical Networks}

Information-theoretical network is a neural network formalism that is
capable of doing associations between two layers of representations.  The
associations can be obtained statistically according to the network's
experiences.

An information-theoretical network has two layers. Each unit of a layer
represents an element in the input or output of a training pattern, which
might be a logogen or a word.  Units in different layers are connected.
The weight of the connection between unit {\it i} in one layer and unit
{\it j} in the other layer is assigned with the mutual information between
the elements represented by the two units

\noindent
(1) \(w_{ij}   =   I(v_{i}, v_{j})   =   \log (Pr(v_{j} v_{i}) /
Pr(v_{i}))\)
\footnote{Where \(v_{i}\) means the event that unit \(i\) is activated.}

Each layer also contains a bias unit, which is always activated.
The weight of the connection between the bias unit in one layer and
unit {\it j} in the other layer is

\noindent
(2)  \(w_{0j}   =   \log Pr(v_{j})\)

Both the information-theoretical network and the back-propagation network
compute the posterior probabilities for an association task
\cite{gorin-89,robinson-92}. However, only the information-theoretical
network is isomorphic to the directly interconnected verbal systems in the
dual-coding theory.  Besides, an information-theoretical network has the
following advantages: (1) it learns fast. The network can learn in a single
pass without gradient decent. (2) it is adaptive. It can incrementally
adapt to new experiences simply by adding new data to the training samples
and modifying the associations according to the changed statistics. These
make the network more psychologically plausible.

\subsection{Lexical Selection as an Associative Process}

We tried to map source language f-structures to target language f-structure
in a connectionist transfer project \cite{wang-94}.  Functionally, there were
two sub-tasks: 1. finding the target sub-structures, their phrasal
categories and their corresponding source structures; 2. finding the head
of a target structure. The second sub-task is a problem of lexical
selection.  It was first implemented with a back-propagation network.

We replaced the back-propagation networks for lexical selection with
information-theoretical networks simulating the associative process in
the dual-coding theory. The networks have two layers of units. Each source
(target) language lexical item is represented by a unit in the input
(output) layer. One network is constructed for each phrasal category (NP, VP,
AP, etc.).

The networks works in the following way: for a target-language f-structure
to be generated, the transfer system knows its phrasal category and its
corresponding source-language f-structure from the networks that perform
the sub-task 1. It then activates the lexical selection network for that
phrasal category with the input units that correspond to the heads of the
source language f-structure and its sub-structures. Through the connections
between the two layers, the output units are activated, and the lexical
item that corresponds to the most active output unit is selected as the
head of the target f-structure. The following example illustrates how the
system selects the head {\em anmelden} for the German XCOMP sub-structure when
it does the transfer from

{\em $[_{\em sentence}$ $[_{\em subj}$ I$]$ would $[_{\em xcomp}$ $[_{\em
subj}$ I$]$ like $[_{\em xcomp}$ $[_{\em subj}$ I$]$ register $[_{\em
pp-adj}$ for the conference$]]]]$ } to

{\em $[_{\em sentence}$ $[_{\em subj}$ Ich$]$ werde $[_{\em xcomp}$ $[_{\em
subj}$ Ich$]$ $[_{\em adj}$ gerne$]$} anmelden $[_{\em pp-adj}$ {\em fuer
der Konferenz}]]]\footnote{The f-structures are simplified here for the
sake of conciseness.}.

Since the structure networks find that there is a VP sub-structure of XCOMP
in the target structure whose corresponding input structure is {\em $[_{\em
xcomp}$ $[_{\em subj}$ I$]$ to register $[_{\em pp-adj}$ for the
conference$]]]$}, it activates the VP lexical selection network's input
units for {\it I, register} and {\it conference}. By propagating the
activation via the associative connections, the unit for {\it anmelden} is
the most active output. Therefore, {\it anmelden} is chosen as the head of
the {\em xcomp} sub-structure.

\subsection{Preliminary Result}

The domain of our work was the Conference Registration Telephony
Conversations.  The lexicon for the task contained about 500 English and
500 German words. There were 300 English/German f-structure pairs available
from other research tasks \cite{osterholtz-92}.  A separate set of 154
sentential f-structures was used to test the generalization performance of
the system. The testing data was collected for an independent task
\cite{jain-91}.

{}From the 300 sentential f-structure pairs, every German VP sub-structure is
extracted and labeled with its English counterpart. The English
counterpart's head and its immediate sub-structures' heads serve as the
input in a sample of VP association, and the German f-structure's head
become the output of the association.  For the above example, the
association ({\em $[_{input}$ I, register, conference$]$ $[_{output}$
anmelden$]$}) is a sample drawn from the f-structures for the VP network.
The training samples for all the other networks are created in the same
way.

The accuracy of our system with information-theoretical network lexical
selection is lower than the one with back-propagation networks (around 84\%
versus around 92\%) for the training data. However, the generalization
performance on the unseen inputs is better (around 70\% versus around
62\%).  The information-theoretical networks do not over-learn as the
back-propagation networks. This is partially due to the reduced number of
free parameters in the information-theoretical networks.

\section{Summary}

The lexical selection approach discussed here has two advantages. First, it
is learnable. Little human effort on knowledge engineering is
required. Secondly, it is psycholinguistically well-founded in that the
approach adopts a local activation processing model instead of relies upon
symbol passing, as symbolic systems usually do.

\end{document}